\documentclass[11pt,a4paper]{article}
\usepackage[hyperref]{emnlp-ijcnlp-2019}
\usepackage{times}
\usepackage{latexsym}
\usepackage{amsmath}
\usepackage{url}
\usepackage{mathdots}
\usepackage{yhmath}
\usepackage{cancel}
\usepackage{color}
\usepackage{siunitx}
\usepackage{array}
\usepackage{multirow}
\usepackage{amssymb}
\usepackage{gensymb}
\usepackage{tabularx}
\usepackage{booktabs}
\usepackage{svg}
\usepackage{tablefootnote}
\usepackage{cleveref}
\usepackage{tabularx}
\usepackage{xcolor}
\usepackage{makecell}
\usepackage{enumitem}

\definecolor{light-gray}{gray}{0.85}

\aclfinalcopy 

\title{
Generating Personalized Recipes from Historical User Preferences
}

\author{Bodhisattwa Prasad Majumder\thanks{\quad denotes equal contribution}, Shuyang Li\footnotemark[1], Jianmo Ni, Julian McAuley \\
 Computer Science and Engineering \\
 University of California, San Diego \\
 {\tt \{bmajumde, shl008, jin018, jmcauley\}@ucsd.edu} \\}

\date{}

\begin{document}
\maketitle
\begin{abstract}
    Existing approaches to recipe generation are unable to create recipes for users with culinary preferences but incomplete knowledge of ingredients in specific dishes.
    We propose a new task of \textit{personalized recipe generation} to help these users: expanding a name and incomplete ingredient details into complete natural-text instructions aligned with the user's historical preferences.
    We attend on technique- and recipe-level representations of a user's previously consumed recipes, fusing these `user-aware' representations in an attention fusion layer to control recipe text generation.
    Experiments on a new dataset of 180K recipes and 700K interactions show our model's ability to generate plausible and personalized recipes compared to non-personalized baselines.
\end{abstract}
\section{Introduction}

In the kitchen, we increasingly rely on instructions from cooking websites: recipes.
A cook with a predilection for Asian cuisine may wish to prepare chicken curry, but may not know all necessary ingredients apart from a few basics.
These users with limited knowledge cannot rely on existing recipe generation approaches that focus on creating coherent recipes given all ingredients and a recipe name \cite{kiddon2016globally}.
Such models do not address issues of personal preference (e.g.~culinary tastes, garnish choices) and incomplete recipe details.
We propose to approach both problems via \textit{personalized generation} of plausible, user-specific recipes using user preferences extracted from previously consumed recipes.

Our work combines two important tasks from natural language processing and recommender systems: data-to-text generation \cite{DBLP:journals/jair/GattK18} and personalized recommendation \cite{DBLP:conf/iui/RashidACLMKR02}.
Our model takes as user input the name of a specific dish, a few key ingredients, and a calorie level.
We pass these loose input specifications to an encoder-decoder framework and attend on user profiles---learned latent representations of recipes previously consumed by a user---to generate a recipe \textit{personalized} to the user's tastes.
We fuse these `user-aware' representations with decoder output in an attention fusion layer to jointly determine text generation.
Quantitative (perplexity, user-ranking) and qualitative analysis on user-aware model outputs confirm that personalization indeed assists in generating plausible recipes from incomplete ingredients.

While personalized text generation has seen success in conveying user writing styles in the product review \cite{DBLP:conf/ijcnlp/NiLVM17, DBLP:conf/acl/NiM18} and dialogue \cite{DBLP:conf/acl/KielaWZDUS18} spaces, we are the first to consider it for the problem of recipe generation, where
output quality is heavily dependent on the \textit{content} of the instructions---such as ingredients and cooking techniques.

To summarize, our main contributions are as follows:
\begin{enumerate}[itemsep=-1ex]
    \item We explore a new task of generating plausible and personalized recipes from incomplete input specifications by leveraging historical user preferences;\footnote{Our source code and appendix are at \url{https://github.com/majumderb/recipe-personalization}}
    \item We release a new dataset of 180K+ recipes and 700K+ user reviews for this task;
    \item We introduce new evaluation strategies for generation quality in instructional texts, centering on quantitative measures of coherence. We also show qualitatively and quantitatively that personalized models generate high-quality and specific recipes that align with historical user preferences.
\end{enumerate}

\section{Related Work}

Large-scale transformer-based language models have shown surprising expressivity and fluency in creative and conditional long-text generation \cite{DBLP:conf/nips/VaswaniSPUJGKP17, radford2019language}.
Recent works have proposed hierarchical methods that condition on narrative frameworks to generate internally consistent long texts \cite{DBLP:conf/acl/LewisDF18, DBLP:conf/emnlp/XuRZZC018, DBLP:journals/corr/abs-1811-05701}.
Here, we generate procedurally structured recipes instead of free-form narratives.

Recipe generation belongs to the field of data-to-text natural language generation \cite{DBLP:journals/jair/GattK18}, which sees other applications in automated journalism \cite{DBLP:conf/inlg/LeppanenMGT17}, question-answering \cite{Antol_2015_ICCV}, and abstractive summarization \cite{DBLP:conf/iclr/PaulusXS18}, among others.
\citet{DBLP:conf/emnlp/KiddonPZC15, DBLP:conf/iclr/BosselutLHEFC18} model recipes as a structured collection of ingredient entities acted upon by cooking actions.
\citet{kiddon2016globally} imposes a `checklist' attention constraint emphasizing hitherto unused ingredients during generation.
\citet{DBLP:conf/emnlp/YangBDL17} attend over explicit ingredient references in the prior recipe step.
Similar hierarchical approaches that infer a full ingredient list to constrain generation will not help personalize recipes, and would be infeasible in our setting due to the potentially unconstrained number of ingredients (from a space of 10K+) in a recipe.
We instead learn historical preferences to guide full recipe generation.

A recent line of work has explored user- and item-dependent aspect-aware review generation \cite{DBLP:conf/ijcnlp/NiLVM17,DBLP:conf/acl/NiM18}.
This work is related to ours in that it combines contextual language generation with personalization.
Here, we attend over historical user preferences from previously consumed recipes to generate recipe content, rather than writing styles.

\section{Approach}
Our model's input specification consists of: the recipe name as a sequence of tokens, a partial list of ingredients, and a caloric level (high, medium, low).
It outputs the recipe instructions as a token sequence: $\mathcal{W}_r=\{w_{r,0}, \dots, w_{r,T}\}$ for a recipe $r$ of length $T$.
To personalize output, we use historical recipe interactions of a user $u \in \mathcal{U}$.

\noindent
\textbf{Encoder:}
Our encoder has three embedding layers: vocabulary embedding $\mathcal{V}$, ingredient embedding $\mathcal{I}$, and caloric-level embedding $\mathcal{C}$.
Each token in the (length $L_n$) recipe name is embedded via $\mathcal{V}$; the embedded token sequence is passed to a two-layered bidirectional GRU (BiGRU) \cite{DBLP:conf/emnlp/ChoMGBBSB14}, which outputs hidden states for names $\{\mathbf{n}_{\text{enc},j} \in \mathbb{R}^{2d_h}\}$, with hidden size $d_h$.
Similarly each of the $L_i$ input ingredients is embedded via $\mathcal{I}$, and the embedded ingredient sequence is passed to another two-layered BiGRU to output ingredient hidden states as $\{\mathbf{i}_{\text{enc},j} \in \mathbb{R}^{2d_h}\}$.
The caloric level is embedded via $\mathcal{C}$ and passed through a projection layer with  weights $W_c$ to generate calorie hidden representation $\mathbf{c}_{\text{enc}} \in \mathbb{R}^{2d_h}$.

\noindent
\textbf{Ingredient Attention:}
We apply attention \cite{bahdanau2014neural} over the encoded ingredients to use encoder outputs at each decoding time step.
We define an attention-score function $\alpha$ with key $K$ and query $Q$: 
\begin{align*}
    \alpha(K, Q) =\frac{\exp \left(\tanh \left(W_{\alpha}\left[K + Q\right]+\mathbf{b}_{\alpha}\right)\right)}{Z} ,
\end{align*}
with trainable weights $W_{\alpha}$, bias $\mathbf{b}_{\alpha}$, and normalization term $Z$.
At decoding time $t$, we calculate the ingredient context $\mathbf{a}_{t}^{i} \in \mathbb{R}^{d_h}$ as:
\begin{align*}
    & \mathbf{a}_{t}^{i}=\sum_{j=1}^{L_i} \alpha\left(\mathbf{i}_{\text{enc},j}, \mathbf{h}_{t}\right) \times \mathbf{i}_{\text{enc},j}.
\end{align*}

\noindent
\textbf{Decoder:}
The decoder is a two-layer GRU with hidden state $h_t$ conditioned on previous hidden state $h_{t-1}$ and input token $w_{r, t}$ from the original recipe text. 
We project the concatenated encoder outputs as the initial decoder hidden state:
\begin{align*}
    & \mathbf{h}_{0} \left(\in \mathbb{R}^{d_h}\right) = W_{h_0} \left[\mathbf{n}_{\text{enc},L_n}; \mathbf{i}_{\text{enc},L_i}; \mathbf{c}_{\text{enc}}\right] + \mathbf{b}_{h_0} \\
    & \mathbf{h}_{t}, \mathbf{o}_{t} = \text{GRU} \left(\left[w_{r, t}; \mathbf{a}_{t}^{i}\right], \mathbf{h}_{t-1} \right).
\end{align*}

To bias  generation toward user preferences, we attend over a user's previously reviewed recipes to jointly determine the final output token distribution.
We consider two different schemes to model preferences from user histories: (1) recipe interactions, and (2) techniques seen therein (defined in \Cref{data}).
\citet{DBLP:conf/uai/RendleFGS09, DBLP:conf/um/QuadranaCJ18, Ueda:2011:UFP:2887675.2887686} explore similar schemes for personalized recommendation.

\begin{figure}[t]
  \centering
  \includegraphics[width=\linewidth]{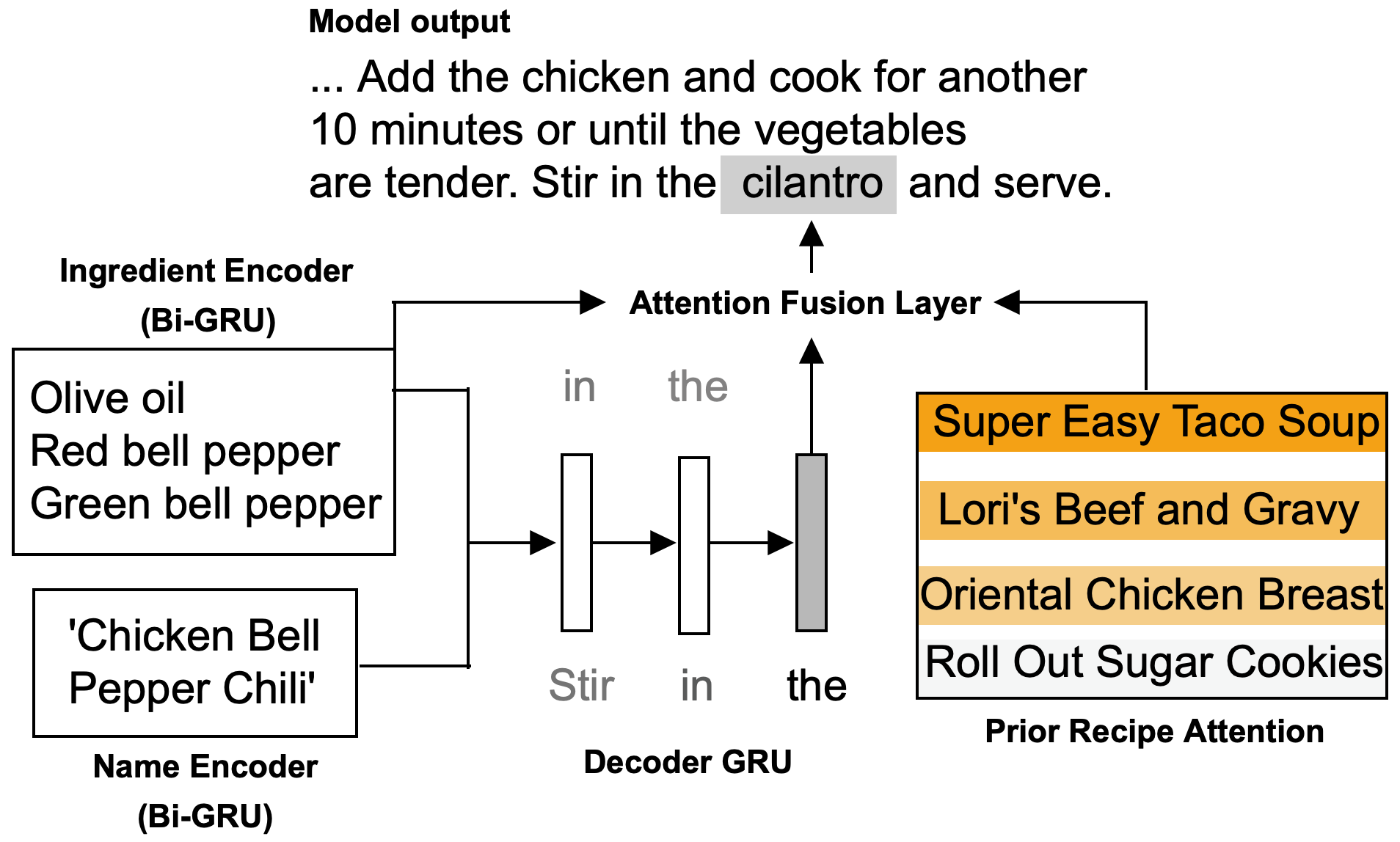}
  \caption{Sample data flow through model architecture. Emphasis on prior recipe attention scores (darker is stronger). Ingredient attention omitted for clarity.}
  \label{fig:ex}
\end{figure}

\noindent
\textbf{Prior Recipe Attention:}
We obtain the set of prior recipes for a user $u$: $R^+_u$, where each recipe can be represented by an embedding from a recipe embedding layer $\mathcal{R}$ or an average of the name tokens embedded by $\mathcal{V}$.
We attend over the $k$-most recent prior recipes, $R^{k+}_u$,  to account for temporal drift of user preferences \cite{DBLP:conf/ismir/MooreCTJ13}.
These embeddings are used in the `\textbf{Prior Recipe}' and `\textbf{Prior Name}' models, respectively.

Given a recipe representation $\mathbf{r} \in \mathbb{R}^{d_r}$ (where $d_r$ is recipe- or vocabulary-embedding size depending on the recipe representation) the \textit{prior recipe attention} context $\mathbf{a}_{t}^{r_u}$ is calculated as
\begin{align*}
    & \mathbf{a}_{t}^{r_u}=\sum_{r \in R^{k+}_u}\alpha\left(\mathbf{r}, \mathbf{h}_{t}\right) \times \mathbf{r}.
\end{align*}

\noindent
\textbf{Prior Technique Attention:}
We calculate prior technique preference (used in the `\textbf{Prior Tech}` model) by normalizing co-occurrence between users and techniques seen in $R^+_u$, to obtain a preference vector $\boldsymbol{\rho}_{u}$.
Each technique $x$ is embedded via a technique embedding layer $\mathcal{X}$ to $\mathbf{x}\in \mathbb{R}^{d_x}$.
\textit{Prior technique attention} is calculated as
\begin{align*}
    & \mathbf{a}_{t}^{x_u}=\sum_{x \text{ seen in } R^+_u} \left(\alpha\left(\mathbf{x}, \mathbf{h}_{t}\right) + \rho_{u, x}\right) \times \mathbf{x},
\end{align*}
where, inspired by copy mechanisms \cite{see2017get, DBLP:conf/acl/GuLLL16}, we add $\rho_{u,x}$ for technique $x$ to emphasize the attention by the user's prior technique preference.

\noindent
\textbf{Attention Fusion Layer:}
We fuse all contexts calculated at time $t$, concatenating them with decoder GRU output and previous token embedding:
\begin{align*}
    & \mathbf{a}^{f}_{t}\!=\!\text{ReLU} \left(W_{f}\!\left[w_{r, t}; \mathbf{o}_{t}; \mathbf{a}_{t}^{i}; (\mathbf{a}_{t}^{r_u} \text{ or } \mathbf{a}_{t}^{x_u})\right]\!+\!\mathbf{b}_f\right)\!.
\end{align*}
We then calculate the token probability:
\begin{align*}
    & P(S_{r, t}) = \text{softmax}\left(W_{P}[\mathbf{a}^f_t] + \mathbf{b}_{P}\right),
\end{align*}
and maximize the log-likelihood of the generated sequence conditioned on input specifications and user preferences.
\Cref{fig:ex} shows a case where the Prior Name model attends strongly on previously consumed savory recipes to suggest the usage of an additional ingredient (`cilantro').

\section{Recipe Dataset: Food.com}
\label{data}
We collect a novel dataset of 230K+ recipe texts and 1M+ user interactions (reviews) over 18 years (2000-2018) from Food.com.\footnote{\url{https://www.kaggle.com/shuyangli94/food-com-recipes-and-user-interactions}}
Here, we restrict to recipes with at least 3 steps, and at least 4 and no more than 20 ingredients. 
We discard users with fewer than 4 reviews, giving 180K+ recipes and 700K+ reviews, with splits as in  \Cref{tab:recipe_ixn_stats}.
\begin{table}[t!]
\small
\centering
\begin{tabular}{@{}lcccc@{}}
\toprule
\bf Split & \bf \# Users & \bf \# Recipes & \bf \# Actions & \bf Sparsity\tablefootnote{Ratio of unobserved actions to all possible actions.} \\ \midrule
Train & 25,076   & 160,901    & 698,901    & 99.983\% \\
Dev   & 7,023    & 6,621      & 7,023      & --       \\
Test  & 12,455   & 11,695     & 12,455     & --       \\ \bottomrule
\end{tabular}
\caption{Statistics of Food.com interactions}
\label{tab:recipe_ixn_stats}
\end{table}
\begin{table*}[t!]
\small
\centering
\begin{tabular}{@{}lcccccc|cc|c@{}}
\toprule
\bf Model      & \bf BPE PPL        & \bf BLEU-1          & \bf BLEU-4         & \bf ROUGE-L         & \bf D-1 (\%) & \bf D-2 (\%) & \bf UMA            & \bf MRR    & \bf PP (\%)            \\ \midrule
NN        & --              & 20.279          & 0.465               &  16.871               & 0.931           & 9.394           & 0.100          & 0.293          & --\\
Enc-Dec    & 9.611          & 28.391          & \textbf{3.385} & \textbf{25.001} & 0.220           & 1.928           & 0.100          & 0.293          & --\\ \midrule
Prior Tech  & 9.572          & \textbf{28.864} & 3.312          & 24.920          & 0.233           & \textbf{2.158}  & 0.128          & 0.319          & 62.821\\
Prior Recipe & 9.551          & 27.858          & 3.215          & 24.822          & 0.231           & 2.062           & 0.302          & 0.412          & \textbf{66.026}\\
Prior Name      & \textbf{9.516} & 28.046          & 3.211          & 24.794          & \textbf{0.233}  & 2.080           & \textbf{0.505} & \textbf{0.628} & 61.165\\ \bottomrule
\end{tabular}
\caption{Metrics on generated recipes from test set. D-1/2 = Distinct-1/2, UMA = User Matching Accuracy, MRR = Mean Reciprocal Rank, PP = Pairwise preference over baseline (evaluated for 310 recipe pairs per model).}
\label{tab:metrics_on_test}
\end{table*}

\begin{table*}[h!]
\small
\centering
\begin{tabularx}{\linewidth}{ l X }
\toprule
\bf Input                   & \textbf{Name:} Pomberrytini; \textbf{Ingredients:} pomegranate-blueberry juice, cranberry juice, \colorbox{light-gray}{vodka}; \textbf{Calorie:} Low \\ \midrule
\bf Gold                    & Place everything except the orange slices in a cocktail shaker. Shake until well mixed and well chilled. Pour into martini glasses and float an orange slice in each glass. \\ \midrule
\bf Enc-Dec                 & Combine all ingredients. Cover and refrigerate. Serve with whipped topping.   \\ \midrule
\bf Prior Tech        & Combine all ingredients. Store in refrigerator. Serve over ice. Enjoy!   \\ \midrule
\bf Prior Recipe   & Pour the ice into a cocktail shaker. Pour in the \colorbox{light-gray}{vodka} and vodka. Add a little water and shake to mix. Pour into the glass and garnish with a slice of orange slices. Enjoy!  \\ \midrule
\bf Prior Name     & Combine all ingredients except for the ice in a blender or food processor. Process to make a smooth paste and then add the remaining \colorbox{light-gray}{vodka} and blend until smooth. Pour into a chilled glass and \colorbox{light-gray}{garnish with a little lemon and fresh mint.}
 \\\bottomrule
\end{tabularx}
\caption{Sample generated recipe. Emphasis on personalization and explicit ingredient mentions via \colorbox{light-gray}{highlights.}}
\label{tab:sample_recipes}
\end{table*}
Our model must learn to generate from a diverse recipe space: in our training data, the average recipe length is 117 tokens with a maximum of 256.
There are 13K unique ingredients across all recipes.
Rare words dominate the vocabulary: 95\% of words appear $<$100 times, accounting for only 1.65\% of all word usage.
As such, we perform Byte-Pair Encoding (BPE) tokenization
\cite{DBLP:conf/acl/SennrichHB16a, radford2018improving}, giving a training vocabulary of 15K tokens across 19M total mentions.
User profiles are similarly diverse: 50\% of users have consumed $\leq$6 recipes, while 10\% of users have consumed $>$45 recipes.

We order reviews by timestamp, keeping the most recent review for each user as the test set, the second most recent for validation, and the remainder for training (sequential leave-one-out evaluation \cite{DBLP:conf/icdm/KangM18}).
We evaluate only on recipes not in the training set.

We manually construct a list of 58 cooking techniques from 384 cooking actions collected by \citet{DBLP:conf/iclr/BosselutLHEFC18}; the most common techniques (\textit{bake}, \textit{combine}, \textit{pour}, \textit{boil}) account for 36.5\% of technique mentions.
We approximate technique adherence via string match between the recipe text and technique list.

\section{Experiments and Results}
For training and evaluation, we provide our model with the first 3-5 ingredients listed in each recipe.
We decode recipe text via top-$k$ sampling \cite{radford2019language}, finding $k=3$ to produce satisfactory results.
We use a hidden size $d_h=256$ for both the encoder and decoder.
Embedding dimensions for vocabulary, ingredient, recipe, techniques, and caloric level are 300, 10, 50, 50, and 5 (respectively).
For prior recipe attention, we set $k=20$, the 80th \%-ile for the number 
of user interactions.
We use the Adam optimizer \cite{DBLP:journals/corr/KingmaB14} with a learning rate of $10^{-3}$, annealed with a decay rate of 0.9 \cite{DBLP:conf/acl/RuderH18}.
We also use teacher-forcing \cite{DBLP:journals/neco/WilliamsZ89} in all training epochs.

In this work, we investigate how leveraging historical user preferences can improve generation quality over strong baselines in our setting.
We compare our personalized models against two baselines.
The first is a name-based Nearest-Neighbor model (\textbf{NN}).
We initially adapted the Neural Checklist Model of \citet{kiddon2016globally} as a baseline; however, we ultimately use a simple Encoder-Decoder baseline with ingredient attention (\textbf{Enc-Dec}), which provides comparable performance and lower complexity.
All personalized models outperform baseline in BPE perplexity (\Cref{tab:metrics_on_test}) with Prior Name performing the best.
While our models exhibit comparable performance to baseline in BLEU-1/4 and ROUGE-L, we generate more diverse (Distinct-1/2: percentage of distinct unigrams and bigrams) and acceptable recipes.
BLEU and ROUGE are not the most appropriate metrics for generation quality.
A `correct' recipe can be written in many ways with the same main entities (ingredients).
As BLEU-1/4 capture structural information via n-gram matching, they are not correlated with subjective recipe quality.
This mirrors observations from \citet{DBLP:conf/emnlp/BahetiRLD18, DBLP:conf/acl/LewisDF18}.

We observe that personalized models make more diverse recipes than baseline. They thus perform better in BLEU-1 with more key entities (ingredient mentions) present, but worse in BLEU-4, as these recipes are written in a personalized way and deviate from gold on the phrasal level. Similarly, the `Prior Name' model generates more unigram-diverse recipes than other personalized models and obtains a correspondingly lower BLEU-1 score.
\medskip

\noindent
\textbf{Qualitative Analysis: }We present sample outputs for a cocktail recipe in \Cref{tab:sample_recipes}, and additional recipes in the appendix.
Generation quality progressively improves from generic baseline output to a blended cocktail produced by our best performing model.
Models attending over prior recipes explicitly reference ingredients.
The Prior Name model further suggests the addition of lemon and mint,
which are reasonably associated with previously consumed recipes like coconut mousse and pork skewers.
\medskip

\noindent
\textbf{Personalization:} To measure personalization, we evaluate how closely the generated text corresponds to a particular user profile.
We compute the likelihood of generated recipes using identical input specifications but conditioned on ten different user profiles---one `gold' user who consumed the original recipe, and nine randomly generated user profiles.
Following \citet{DBLP:conf/acl/LewisDF18}, we expect the highest likelihood for the recipe conditioned on the gold user.
We measure user matching accuracy (UMA)---the proportion where the gold user is ranked highest---and Mean Reciprocal Rank (MRR) \cite{DBLP:conf/lrec/RadevQWF02} of the gold user.
All personalized models beat baselines in both metrics, showing our models personalize generated recipes to the given user profiles.
The Prior Name model achieves the best UMA and MRR by a large margin, revealing that prior recipe names are strong signals for personalization.
Moreover, the addition of attention mechanisms to capture these signals improves language modeling performance over a strong non-personalized baseline.
\medskip

\noindent
\textbf{Recipe Level Coherence:}
A plausible recipe should possess a coherent step order, and we evaluate this via a metric for recipe-level coherence.
We use the neural scoring model from \citet{DBLP:conf/naacl/BosselutCHGHC18} to measure recipe-level coherence for each generated recipe.
Each recipe step is encoded by BERT \cite{DBLP:conf/naacl/DevlinCLT19}.
Our scoring model is a GRU network that learns the overall recipe step ordering structure by minimizing the cosine similarity of recipe step hidden representations presented in the correct and reverse orders.
Once pretrained, our scorer calculates the similarity of a generated recipe to the forward and backwards ordering of its corresponding gold label, giving a score equal to the difference between the former and latter. A higher score indicates better step ordering (with a maximum score of 2).
\Cref{tab:coherence_metrics} shows that our personalized models achieve average recipe-level coherence scores of 1.78-1.82, surpassing the baseline at 1.77.
\medskip

\noindent
\textbf{Recipe Step Entailment:}
Local coherence is also crucial to a user following a recipe: it is crucial that subsequent steps are logically consistent with prior ones.
We model local coherence as an entailment task: predicting the likelihood that a recipe step follows the preceding.
We sample several consecutive (positive) and non-consecutive (negative) pairs of steps from each recipe.
We train a BERT \cite{DBLP:conf/naacl/DevlinCLT19} model to predict the entailment score of a pair of steps separated by a \texttt{[SEP]} token, using the final representation of the \texttt{[CLS]} token.
The step entailment score is computed as the average of scores for each set of consecutive steps in each recipe, averaged over every generated recipe for a model, as shown in \Cref{tab:coherence_metrics}.
\medskip

\noindent
\textbf{Human Evaluation:}
We presented 310 pairs of recipes for pairwise comparison \cite{DBLP:conf/acl/LewisDF18} (details in appendix) between baseline and each personalized model, with results shown in \Cref{tab:metrics_on_test}.
On average, human evaluators preferred personalized model outputs to baseline 63\% of the time, confirming that personalized attention improves the semantic plausibility of generated recipes.
We also performed a small-scale human coherence survey over 90 recipes, in which 60\% of users found recipes generated by personalized models to be more coherent and preferable to those generated by baseline models. 

\begin{table}[t!]
\small
\centering
\begin{tabular}{@{}lcc@{}}
\toprule
\bf Model       & \bf \thead{Recipe Level\\Coherence}      & \bf \thead{Recipe Step\\ Entailment}  \\ \midrule
Enc-Dec         & 1.77                              & 0.72                                           \\  \midrule
Prior Tech      & 1.78                              & 0.73                                                \\
Prior Recipe    & 1.80                              & 0.76                                             \\
Prior Name      & \textbf{1.82}                     & \textbf{0.78}                                      \\ \bottomrule
\end{tabular}
\caption{Coherence metrics on generated recipes from test set.}
\label{tab:coherence_metrics}
\end{table}

\section{Conclusion}
In this paper, we propose a novel task: to generate personalized recipes from incomplete input specifications and user histories.
On a large novel dataset of 180K recipes and 700K reviews, we show that our personalized generative models can generate plausible, personalized, and coherent recipes preferred by human evaluators for consumption.
We also introduce a set of automatic coherence measures for instructional texts as well as personalization metrics to support our claims.
Our future work includes generating structured representations of recipes to handle ingredient properties, as well as accounting for references to collections of ingredients (e.g. ``dry mix").
\medskip

\noindent\textbf{Acknowledgements.}
This work is partly supported by NSF \#1750063.
We thank all reviewers for their constructive suggestions, as well as Rei M., Sujoy P., Alicia L., Eric H., Tim S., Kathy C., Allen C., and Micah I. for their feedback.

\bibliography{emnlp-ijcnlp-2019}

\begin{thebibliography}{35}
\expandafter\ifx\csname natexlab\endcsname\relax\def\natexlab#1{#1}\fi

\bibitem[{Agrawal et~al.(2017)Agrawal, Lu, Antol, Mitchell, Zitnick, Parikh,
  and Batra}]{Antol_2015_ICCV}
Aishwarya Agrawal, Jiasen Lu, Stanislaw Antol, Margaret Mitchell, C.~Lawrence
  Zitnick, Devi Parikh, and Dhruv Batra. 2017.
\newblock \href {https://doi.org/10.1007/s11263-016-0966-6} {{VQA:} visual
  question answering}.
\newblock \emph{IJCV}, 123(1):4--31.

\bibitem[{Bahdanau et~al.(2015)Bahdanau, Cho, and Bengio}]{bahdanau2014neural}
Dzmitry Bahdanau, Kyunghyun Cho, and Yoshua Bengio. 2015.
\newblock \href {http://arxiv.org/abs/1409.0473} {Neural machine translation by
  jointly learning to align and translate}.
\newblock In \emph{ICLR}.

\bibitem[{Baheti et~al.(2018)Baheti, Ritter, Li, and
  Dolan}]{DBLP:conf/emnlp/BahetiRLD18}
Ashutosh Baheti, Alan Ritter, Jiwei Li, and Bill Dolan. 2018.
\newblock \href {https://aclanthology.info/papers/D18-1431/d18-1431}
  {Generating more interesting responses in neural conversation models with
  distributional constraints}.
\newblock In \emph{EMNLP}.

\bibitem[{Bosselut et~al.(2018{\natexlab{a}})Bosselut, {\c{C}}elikyilmaz, He,
  Gao, Huang, and Choi}]{DBLP:conf/naacl/BosselutCHGHC18}
Antoine Bosselut, Asli {\c{C}}elikyilmaz, Xiaodong He, Jianfeng Gao, Po{-}Sen
  Huang, and Yejin Choi. 2018{\natexlab{a}}.
\newblock \href {https://aclanthology.info/papers/N18-1016/n18-1016}
  {Discourse-aware neural rewards for coherent text generation}.
\newblock In \emph{NAACL-HLT}.

\bibitem[{Bosselut et~al.(2018{\natexlab{b}})Bosselut, Levy, Holtzman, Ennis,
  Fox, and Choi}]{DBLP:conf/iclr/BosselutLHEFC18}
Antoine Bosselut, Omer Levy, Ari Holtzman, Corin Ennis, Dieter Fox, and Yejin
  Choi. 2018{\natexlab{b}}.
\newblock \href {https://openreview.net/forum?id=rJYFzMZC-} {Simulating action
  dynamics with neural process networks}.
\newblock In \emph{ICLR}.

\bibitem[{Cho et~al.(2014)Cho, van Merrienboer, G{\"{u}}l{\c{c}}ehre, Bahdanau,
  Bougares, Schwenk, and Bengio}]{DBLP:conf/emnlp/ChoMGBBSB14}
Kyunghyun Cho, Bart van Merrienboer, {\c{C}}aglar G{\"{u}}l{\c{c}}ehre, Dzmitry
  Bahdanau, Fethi Bougares, Holger Schwenk, and Yoshua Bengio. 2014.
\newblock \href {http://aclweb.org/anthology/D/D14/D14-1179.pdf} {Learning
  phrase representations using {RNN} encoder-decoder for statistical machine
  translation}.
\newblock In \emph{EMNLP}.

\bibitem[{Devlin et~al.(2019)Devlin, Chang, Lee, and
  Toutanova}]{DBLP:conf/naacl/DevlinCLT19}
Jacob Devlin, Ming{-}Wei Chang, Kenton Lee, and Kristina Toutanova. 2019.
\newblock \href {https://aclweb.org/anthology/papers/N/N19/N19-1423/} {{BERT:}
  pre-training of deep bidirectional transformers for language understanding}.
\newblock In \emph{NAACL-HLT 2019}.

\bibitem[{Fan et~al.(2018)Fan, Lewis, and Dauphin}]{DBLP:conf/acl/LewisDF18}
Angela Fan, Mike Lewis, and Yann Dauphin. 2018.
\newblock \href {https://aclanthology.info/papers/P18-1082/p18-1082}
  {Hierarchical neural story generation}.
\newblock In \emph{ACL}.

\bibitem[{Gatt and Krahmer(2018)}]{DBLP:journals/jair/GattK18}
Albert Gatt and Emiel Krahmer. 2018.
\newblock \href {https://doi.org/10.1613/jair.5477} {Survey of the state of the
  art in natural language generation: Core tasks, applications and evaluation}.
\newblock \emph{J. Artif. Intell. Res.}, 61:65--170.

\bibitem[{Gu et~al.(2016)Gu, Lu, Li, and Li}]{DBLP:conf/acl/GuLLL16}
Jiatao Gu, Zhengdong Lu, Hang Li, and Victor O.~K. Li. 2016.
\newblock \href {http://aclweb.org/anthology/P/P16/P16-1154.pdf} {Incorporating
  copying mechanism in sequence-to-sequence learning}.
\newblock In \emph{ACL}.

\bibitem[{Howard and Ruder(2018)}]{DBLP:conf/acl/RuderH18}
Jeremy Howard and Sebastian Ruder. 2018.
\newblock \href {https://doi.org/10.18653/v1/P18-1031} {Universal language
  model fine-tuning for text classification}.
\newblock In \emph{ACL}.

\bibitem[{Kang and McAuley(2018)}]{DBLP:conf/icdm/KangM18}
Wang{-}Cheng Kang and Julian McAuley. 2018.
\newblock \href {https://doi.org/10.1109/ICDM.2018.00035} {Self-attentive
  sequential recommendation}.
\newblock In \emph{ICDM}.

\bibitem[{Kiddon et~al.(2015)Kiddon, Ponnuraj, Zettlemoyer, and
  Choi}]{DBLP:conf/emnlp/KiddonPZC15}
Chlo{\'{e}} Kiddon, Ganesa~Thandavam Ponnuraj, Luke Zettlemoyer, and Yejin
  Choi. 2015.
\newblock \href {http://aclweb.org/anthology/D/D15/D15-1114.pdf} {Mise en
  place: Unsupervised interpretation of instructional recipes}.
\newblock In \emph{EMNLP}.

\bibitem[{Kiddon et~al.(2016)Kiddon, Zettlemoyer, and
  Choi}]{kiddon2016globally}
Chlo{\'{e}} Kiddon, Luke Zettlemoyer, and Yejin Choi. 2016.
\newblock \href {http://aclweb.org/anthology/D/D16/D16-1032.pdf} {Globally
  coherent text generation with neural checklist models}.
\newblock In \emph{EMNLP}.

\bibitem[{Kingma and Ba(2015)}]{DBLP:journals/corr/KingmaB14}
Diederik~P. Kingma and Jimmy Ba. 2015.
\newblock \href {http://arxiv.org/abs/1412.6980} {Adam: {A} method for
  stochastic optimization}.
\newblock In \emph{ICLR}.

\bibitem[{Lepp{\"{a}}nen et~al.(2017)Lepp{\"{a}}nen, Munezero,
  Granroth{-}Wilding, and Toivonen}]{DBLP:conf/inlg/LeppanenMGT17}
Leo Lepp{\"{a}}nen, Myriam Munezero, Mark Granroth{-}Wilding, and Hannu
  Toivonen. 2017.
\newblock \href {https://aclanthology.info/papers/W17-3528/w17-3528}
  {Data-driven news generation for automated journalism}.
\newblock In \emph{INLG}.

\bibitem[{Moore et~al.(2013)Moore, Chen, Turnbull, and
  Joachims}]{DBLP:conf/ismir/MooreCTJ13}
Joshua~L. Moore, Shuo Chen, Douglas Turnbull, and Thorsten Joachims. 2013.
\newblock \href
  {http://www.ppgia.pucpr.br/ismir2013/wp-content/uploads/2013/09/220\_Paper.pdf}
  {Taste over time: The temporal dynamics of user preferences}.
\newblock In \emph{ISMIR}.

\bibitem[{Ni et~al.(2017)Ni, Lipton, Vikram, and
  McAuley}]{DBLP:conf/ijcnlp/NiLVM17}
Jianmo Ni, Zachary~C. Lipton, Sharad Vikram, and Julian McAuley. 2017.
\newblock \href {https://aclanthology.info/papers/I17-1079/i17-1079}
  {Estimating reactions and recommending products with generative models of
  reviews}.
\newblock In \emph{IJCNLP}.

\bibitem[{Ni and McAuley(2018)}]{DBLP:conf/acl/NiM18}
Jianmo Ni and Julian McAuley. 2018.
\newblock \href {https://aclanthology.info/papers/P18-2112/p18-2112}
  {Personalized review generation by expanding phrases and attending on
  aspect-aware representations}.
\newblock In \emph{ACL}.

\bibitem[{Paulus et~al.(2018)Paulus, Xiong, and
  Socher}]{DBLP:conf/iclr/PaulusXS18}
Romain Paulus, Caiming Xiong, and Richard Socher. 2018.
\newblock \href {https://openreview.net/forum?id=HkAClQgA-} {A deep reinforced
  model for abstractive summarization}.
\newblock In \emph{ICLR}.

\bibitem[{Quadrana et~al.(2018)Quadrana, Cremonesi, and
  Jannach}]{DBLP:conf/um/QuadranaCJ18}
Massimo Quadrana, Paolo Cremonesi, and Dietmar Jannach. 2018.
\newblock \href {https://doi.org/10.1145/3209219.3209270} {Sequence-aware
  recommender systems}.
\newblock In \emph{UMAP}.

\bibitem[{Radev et~al.(2002)Radev, Qi, Wu, and Fan}]{DBLP:conf/lrec/RadevQWF02}
Dragomir~R. Radev, Hong Qi, Harris Wu, and Weiguo Fan. 2002.
\newblock \href {http://www.lrec-conf.org/proceedings/lrec2002/pdf/301.pdf}
  {Evaluating web-based question answering systems}.
\newblock In \emph{LREC}.

\bibitem[{Radford et~al.(2018)Radford, Narasimhan, Salimans, and
  Sutskever}]{radford2018improving}
Alec Radford, Karthik Narasimhan, Tim Salimans, and Ilya Sutskever. 2018.
\newblock \href
  {https://s3-us-west-2.amazonaws.com/openai-assets/research-covers/language-unsupervised/language_understanding_paper.pdf}
  {Improving language understanding by generative pre-training}.

\bibitem[{Radford et~al.(2019)Radford, Wu, Child, Luan, Amodei, and
  Sutskever}]{radford2019language}
Alec Radford, Jeff Wu, Rewon Child, David Luan, Dario Amodei, and Ilya
  Sutskever. 2019.
\newblock \href {https://openai.com/blog/better-language-models/} {Language
  models are unsupervised multitask learners}.

\bibitem[{Rashid et~al.(2002)Rashid, Albert, Cosley, Lam, McNee, Konstan, and
  Riedl}]{DBLP:conf/iui/RashidACLMKR02}
Al~Mamunur Rashid, Istvan Albert, Dan Cosley, Shyong~K. Lam, Sean~M. McNee,
  Joseph~A. Konstan, and John Riedl. 2002.
\newblock \href {https://doi.org/10.1145/502716.502737} {Getting to know you:
  learning new user preferences in recommender systems}.
\newblock In \emph{IUI}.

\bibitem[{Rendle et~al.(2009)Rendle, Freudenthaler, Gantner, and
  Schmidt{-}Thieme}]{DBLP:conf/uai/RendleFGS09}
Steffen Rendle, Christoph Freudenthaler, Zeno Gantner, and Lars
  Schmidt{-}Thieme. 2009.
\newblock \href
  {https://dslpitt.org/uai/displayArticleDetails.jsp?mmnu=1\&smnu=2\&article\_id=1630\&proceeding\_id=25}
  {{BPR:} bayesian personalized ranking from implicit feedback}.
\newblock In \emph{UAI}.

\bibitem[{See et~al.(2017)See, Liu, and Manning}]{see2017get}
Abigail See, Peter~J. Liu, and Christopher~D. Manning. 2017.
\newblock \href {https://doi.org/10.18653/v1/P17-1099} {Get to the point:
  Summarization with pointer-generator networks}.
\newblock In \emph{ACL}.

\bibitem[{Sennrich et~al.(2016)Sennrich, Haddow, and
  Birch}]{DBLP:conf/acl/SennrichHB16a}
Rico Sennrich, Barry Haddow, and Alexandra Birch. 2016.
\newblock \href {http://aclweb.org/anthology/P/P16/P16-1162.pdf} {Neural
  machine translation of rare words with subword units}.
\newblock In \emph{ACL}.

\bibitem[{Ueda et~al.(2011)Ueda, Takahata, and
  Nakajima}]{Ueda:2011:UFP:2887675.2887686}
Mayumi Ueda, Mari Takahata, and Shinsuke Nakajima. 2011.
\newblock \href {http://dl.acm.org/citation.cfm?id=2887675.2887686} {User's
  food preference extraction for personalized cooking recipe recommendation}.
\newblock In \emph{SPIM}.

\bibitem[{Vaswani et~al.(2017)Vaswani, Shazeer, Parmar, Uszkoreit, Jones,
  Gomez, Kaiser, and Polosukhin}]{DBLP:conf/nips/VaswaniSPUJGKP17}
Ashish Vaswani, Noam Shazeer, Niki Parmar, Jakob Uszkoreit, Llion Jones,
  Aidan~N. Gomez, Lukasz Kaiser, and Illia Polosukhin. 2017.
\newblock \href {http://papers.nips.cc/paper/7181-attention-is-all-you-need}
  {Attention is all you need}.
\newblock In \emph{NIPS}.

\bibitem[{Williams and Zipser(1989)}]{DBLP:journals/neco/WilliamsZ89}
Ronald~J. Williams and David Zipser. 1989.
\newblock \href {https://doi.org/10.1162/neco.1989.1.2.270} {A learning
  algorithm for continually running fully recurrent neural networks}.
\newblock \emph{Neural Computation}, 1(2):270--280.

\bibitem[{Xu et~al.(2018)Xu, Ren, Zhang, Zeng, Cai, and
  Sun}]{DBLP:conf/emnlp/XuRZZC018}
Jingjing Xu, Xuancheng Ren, Yi~Zhang, Qi~Zeng, Xiaoyan Cai, and Xu~Sun. 2018.
\newblock \href {https://aclanthology.info/papers/D18-1462/d18-1462} {A
  skeleton-based model for promoting coherence among sentences in narrative
  story generation}.
\newblock In \emph{EMNLP}.

\bibitem[{Yang et~al.(2017)Yang, Blunsom, Dyer, and
  Ling}]{DBLP:conf/emnlp/YangBDL17}
Zichao Yang, Phil Blunsom, Chris Dyer, and Wang Ling. 2017.
\newblock \href {https://aclanthology.info/papers/D17-1197/d17-1197}
  {Reference-aware language models}.
\newblock In \emph{EMNLP}.

\bibitem[{Yao et~al.(2018)Yao, Peng, Weischedel, Knight, Zhao, and
  Yan}]{DBLP:journals/corr/abs-1811-05701}
Lili Yao, Nanyun Peng, Ralph~M. Weischedel, Kevin Knight, Dongyan Zhao, and Rui
  Yan. 2018.
\newblock \href {http://arxiv.org/abs/1811.05701} {Plan-and-write: Towards
  better automatic storytelling}.
\newblock \emph{CoRR}, abs/1811.05701.

\bibitem[{Zhang et~al.(2018)Zhang, Dinan, Urbanek, Szlam, Kiela, and
  Weston}]{DBLP:conf/acl/KielaWZDUS18}
Saizheng Zhang, Emily Dinan, Jack Urbanek, Arthur Szlam, Douwe Kiela, and Jason
  Weston. 2018.
\newblock \href {https://aclanthology.info/papers/P18-1205/p18-1205}
  {Personalizing dialogue agents: {I} have a dog, do you have pets too?}
\newblock In \emph{ACL}.

\end{thebibliography}
\bibliographystyle{acl_natbib}

\clearpage

\section*{Appendix}

\subsection*{Food.com: Dataset Details}
Our raw data consists of 270K recipes and 1.4M user-recipe interactions (reviews) scraped from Food.com, covering a period of 18 years (January 2000 to December 2018).
See \Cref{tab:int-stats} for dataset summary statistics, and \Cref{tab:sample_gk} for sample information about one user-recipe interaction and the recipe involved.

\begin{table}[h!]
\small
\centering
\begin{tabular}{@{}lcccc@{}}
\toprule
          & \# Recipes & \# Users & \# Reviews & Sparsity (\%) \\ \midrule
Raw       & 231,637    & 226,570  & 1,132,367  & 99.998        \\
Processed & 178,265    & 25,076   & 749,053    & 99.983        \\ \bottomrule
\end{tabular}
\caption{Interaction statistics for Food.com dataset before and after data processing.}
\label{tab:int-stats}
\end{table}

\begin{table*}[h!]
\centering
\begin{tabular}{@{}lc@{}}
\toprule
Field          & Value  \\ \midrule
date           & 2002-03-30   \\
user\_id       & 27395 \\
recipe\_id     & 23933  \\
name           & chinese candy \\
n\_steps       & 4  \\
steps          & \begin{tabular}[c]{@{}c@{}}{[}`melt butterscotch chips in heavy saucepan over low heat',\\  `fold in peanuts and chinese noodles until coated',\\  'drop by tablespoon onto waxed paper',\\  `let stand in cool place until firm'{]}\end{tabular} \\
n\_ingredients & 3 \\
ingredients    & {[}`butterscotch chips', `chinese noodles', `salted peanuts'{]} \\
calorie\_level & 0 (Low)  \\ \bottomrule
\end{tabular}
\caption{Sample data from GeniusKitchen with recipe and user interaction details.}
\label{tab:sample_gk}
\end{table*}

\subsection*{Generated Examples}
See \Cref{tab:sample_chx} for a sample recipe for chicken chili and \Cref{tab:sample_waffle} for a sample recipe for sweet waffles.

\begin{table*}[h!]
\small
\centering
\begin{tabularx}{\linewidth}{ l X }
\toprule
Input                   & \textbf{Name:} Chicken Bell Pepper Chili Weight Watchers

\textbf{Ingredients:} olive oil, red bell pepper, green bell pepper, yellow bell pepper

\textbf{Calorie:} Medium \\ \midrule
Gold                    & In a large nonstick dutch oven or saucepan, heat the oil.

Saute the bell peppers and onion until softened, about 5 minutes.

Add the chicken.

Cook, breaking apart with a wooden spoon, until no longer pink, 5 - 7 minutes.

Sprinkle with the flour, chili powder, cumin, cocoa powder, and cayenne.

Cook, stirring briskly and constantly, 1 minute. Stir in the tomatoes and vinegar.

Bring to a boil.

Reduce the heat and simmer, stirring frequently, until thick, 30 - 40 minutes.

Stir in the chickpeas.

Simmer until heated through, about 5 minutes. \\ \midrule
Enc-Dec                 & Heat oil in a large, heavy pot over medium heat.

Add chicken and cook until browned on both sides, about 10 minutes, or until browned; drain.

Add onion and garlic and saute for 2 minutes.

Add chicken broth, chicken broth, and corn starch.

Bring to a slow boil, reduce heat, cover and simmer for 30 minutes.

Remove from heat and add the tomatoes and corn.

Simmer until chicken is cooked through, about 10 minutes.

Serve over rice or noodles.   \\ \midrule
Prior Tech        & Heat oil in large saucepan over medium - high heat.

Add chicken and cook until browned.

Add bell pepper and onions, cook 2 minutes, stirring frequently.

Add chicken broth.

Cover, simmer, stirring occasionally, for 10 minutes or until vegetables are tender, stirring occasionally.

Add chicken, cover, and cook 10 more minutes or until chicken is cooked through.

Remove from heat.

Stir in remaining ingredients.

Season with salt and pepper to taste.

Serve over rice and top with additional shredded cheese and cilantro.   \\ \midrule
Prior Recipe   & Heat oil in large dutch oven over medium - high.

Add chicken, bell peppers, bell peppers and onion.

Saute 2 - 3 minutes.

Add broth and tomato paste; stir.

Bring to a boil. 

Reduce heat and simmer, covered,

for 10 minutes or until rice is tender, stirring often.

Stir in chicken.

Cover and simmer for 10 minutes.

Stir in parsley.

Serve over rice.

Garnish with cilantro.  \\ \midrule
Prior Name     & Heat the oil in a large skillet over medium - high heat.

Add the peppers and onions.

Cook, stirring, until the vegetables are soft and beginning to soften, about 5 minutes.

Stir in the tomatoes, corn, corn, and corn.

Bring to a boil.

Reduce heat to low, cover, and simmer for 10 minutes.

Add the chicken and cook for another 10 minutes or until the vegetables are tender.

Stir in the cilantro and serve.
 \\\bottomrule
\end{tabularx}
\caption{Sample generated recipe ``Chicken Bell Pepper Chili Weight Watchers'' for all models.}
\label{tab:sample_chx}
\end{table*}

\begin{table*}[t!]
\begin{tabularx}{\linewidth}{ l X }
\toprule
Input                   & \textbf{Name:} Cinna Nilla Waffles

\textbf{Ingredients:} flour, egg, milk, vegetable oil, sugar

\textbf{Calorie:} Medium \\ \midrule
Gold                    & Heat waffle iron.

Beat eggs in large bowl with hand beater until fluffy.

Beat in remaining ingredients until smooth.

Pour batter into waffle iron and cook for approximately 5 minutes. 

Serve immediately or cool to pack into freezer bags and freeze. \\ \midrule
Enc-Dec                 & Mix the flour, baking soda, and salt in a bowl.

In a separate bowl, whisk together the milk, oil, eggs and sugar.

Stir the flour mixture into the egg mixture, and continue to mix well.

Add the flour to the egg mixture.

Mix well and pour into a greased waffle iron.

Cook for 2 minutes, remove from heat and serve.   \\ \midrule
Prior Tech        & In a medium bowl mix flour, eggs, and milk until combined.

Add the dry ingredients and stir until just combined and do not mix.

Heat griddle over medium heat, add the oil, oil, and cook the pancakes until golden brown and cooked through.

Serve with a little milk or cream. Enjoy   \\ \midrule
Prior Recipe   & In a mixing bowl, whisk together the eggs, milk, oil, sugar, vanilla, salt and vanilla.
                                                              
Cover and let the mixture stand in the fridge for about 1 hour.

Spoon batter into waffle iron and close the grill.  \\ \midrule
Prior Name     & Preheat waffle iron.

Beat together the eggs, milk and oil until well blended, add the vanilla and mix well with a mixer.

Fold in flour, baking powder, and cinnamon.

Spread 1 / 2 the mixture in a greased waffle iron.

Bake until golden brown, about 15 minutes per side.

Sprinkle with powdered sugar and serve warm.
 \\\bottomrule
\end{tabularx}
\caption{Sample generated waffle recipe for all models.}
\label{tab:sample_waffle}
\end{table*}


\section*{Human Evaluation}
\begin{figure*}[t]
  \centering
  \includegraphics[width=\linewidth]{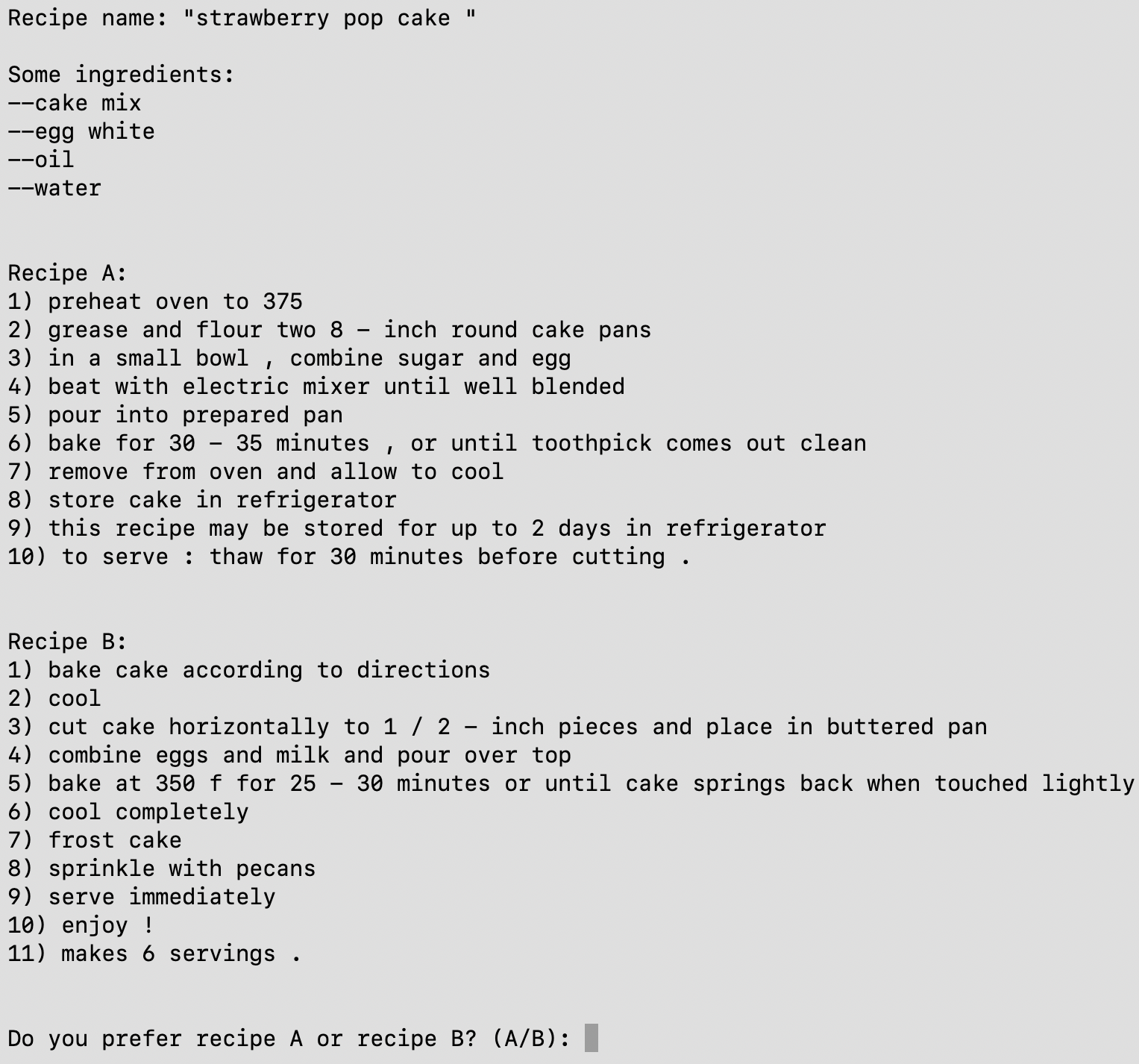}
  \caption{A sample question for pairwise evaluation survey.}
  \label{fig:ex_eval}
\end{figure*}

\begin{figure*}[h!]
  \centering
  \includegraphics[width=\linewidth]{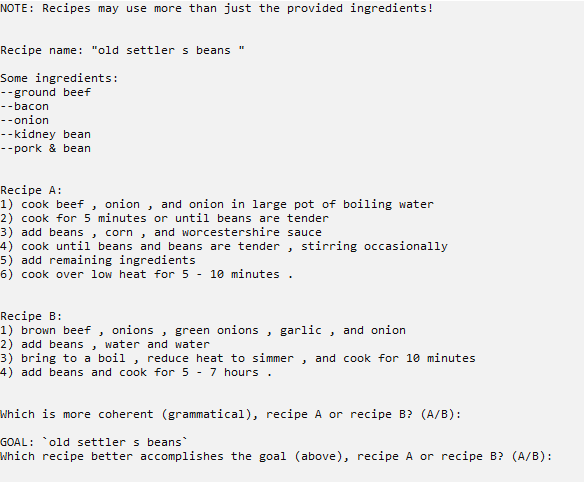}
  \caption{A sample question for coherence evaluation survey.}
  \label{fig:ex_eval2}
\end{figure*}

We prepared a set of 15 pairwise comparisons per evaluation session, and collected 930 pairwise evaluations (310 per personalized model) over 62 sessions.
For each pair, users were given a partial recipe specification (name and 3-5 key ingredients), as well as two generated recipes labeled `A' and `B'.
One recipe is generated from our baseline encoder-decoder model and one recipe is generated by one of our three personalized models (Prior Tech, Prior Name, Prior Recipe).
The order of recipe presentation (A/B) is randomly selected for each question.
A screenshot of the user evaluation interface is given in \Cref{fig:ex_eval}.
We ask the user to indicate which recipe they find more coherent, and which recipe best accomplishes the goal indicated by the recipe name.
A screenshot of this survey interface is given in \Cref{fig:ex_eval2}.


\end{document}